%% file: main.tex
\documentclass[10pt,twocolumn,letterpaper]{article}

\usepackage{iccv}
\usepackage{times}
\usepackage{epsfig}
\usepackage{graphicx}
\usepackage{amsmath}
\usepackage{amssymb}

\usepackage{algorithm}
\usepackage{algorithmicx}  
\usepackage{algpseudocode}
\usepackage{color}
\usepackage{tabularx}
\usepackage{pifont}
\usepackage{multirow}
\usepackage{caption}
\usepackage{subcaption}
\usepackage{graphics}
\usepackage{mwe}
\usepackage{balance}
\usepackage{xcolor}

\usepackage{xr-hyper}
\usepackage[pagebackref=true,breaklinks=true,letterpaper=true,colorlinks,bookmarks=false]{hyperref}

\iccvfinalcopy 

\def\iccvPaperID{3110} 

\ificcvfinal\fi

\begin{document}

\title{DRIVE: Deep Reinforced Accident Anticipation with Visual Explanation}

\author{Wentao Bao, Qi Yu, Yu Kong\\
Rochester Institute of Technology, Rochester, NY 14623, USA\\
{\tt\small \{wb6219, qi.yu, yu.kong\}@rit.edu}
}

\maketitle
\ificcvfinal\thispagestyle{empty}\fi

\begin{abstract}
   Traffic accident anticipation aims to accurately and promptly predict the occurrence of a future accident from dashcam videos, which is vital for a safety-guaranteed self-driving system. To encourage an early and accurate decision, existing approaches typically focus on capturing the cues of spatial and temporal context before a future accident occurs. However, their decision-making lacks visual explanation and ignores the dynamic interaction with the environment. In this paper, we propose Deep ReInforced accident anticipation with Visual Explanation, named DRIVE. The method simulates both the bottom-up and top-down visual attention mechanism in a dashcam observation environment so that the decision from the proposed stochastic multi-task agent can be visually explained by attentive regions. Moreover, the proposed dense anticipation reward and sparse fixation reward are effective in training the DRIVE model with our improved reinforcement learning algorithm. Experimental results show that the DRIVE model achieves state-of-the-art performance on multiple real-world traffic accident datasets. Code and pre-trained model are available at {\small{\url{https://www.rit.edu/actionlab/drive}}}.
\end{abstract}

\section{Introduction}

With increasing demand for autonomous driving, anticipating possible future accidents is becoming the central consideration to guarantee a safe driving strategy~\cite{ChanACCV2016,SuzukiCVPR2018,BaoMM2020}. Given a dashcam video, an accident anticipation model aims to tell the driving system if and when a traffic accident will occur in the near future.
Despite remarkable advances in visual perception~\cite{DengTITS2020,GeigerCVPR2012,HeCVPR2016}, the decision-making of driving control has long been studied in isolation with vision perception research for the autonomous driving scenario~\cite{kimICCV2017,SaxenaICRA2020}. We target at bridging this gap by investigating a key research question: \emph{where do drivers look when predicting possible future accidents?} This will lead to a visually explainable model that associates the low-level visual attention and high-level accident anticipation. 

\begin{figure}
    \centering
    \includegraphics[width=\linewidth]{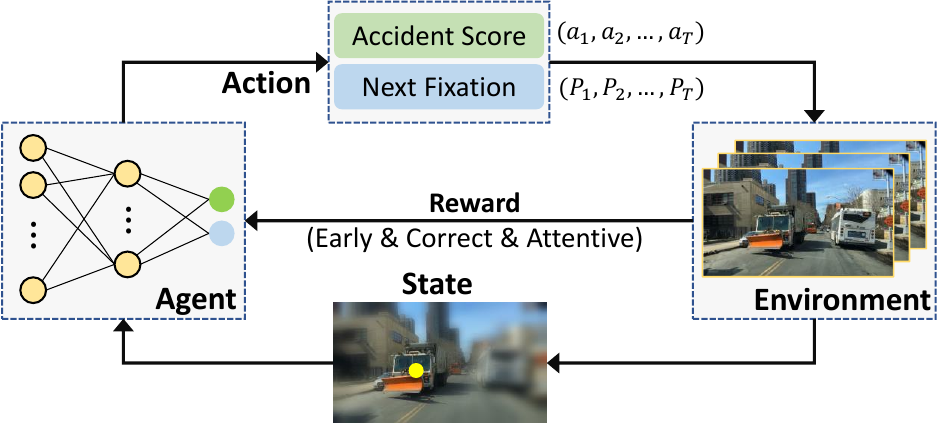}
    \caption{\textbf{The Markov decision process of the DRIVE model.} The neural network agent (left) learns to \emph{exploit} visual attentive state (bottom) to predict the actions including the accident score and the next fixation (top), which in return \emph{explore} the driving environment (right) to maximize the total reward (middle). 
    }
    \label{fig:overview}
\end{figure}

The traffic accident anticipation is far from being solved due to the following challenges. First, the visual cues of a future accident are vital to training a discriminative model but in practice, they are difficult to be captured from the limited and noisy video data before the accident occurs. Previous works take advantage of object detection and learn the accident visual cues by either soft attention in~\cite{ChanACCV2016} or graph relational learning in~\cite{BaoMM2020}. In this paper, we propose to explicitly learn the visual attention behavior to address \emph{where to look} such that accident risky regions can be localized. 

Second, it is intrinsically a trade-off between an \emph{early} decision and a \emph{correct} decision since the earlier to anticipate an accident, the harder to make the decision right due to fewer accident-relevant cues. Existing works~\cite{ChanACCV2016,BaoMM2020} simply address the trade-off by training supervised deep learning models with an exponentially weighted classification loss. In this paper, we address this trade-off by formulating the task as a Markov Decision Process (MDP), where exploration and exploitation can be dynamically balanced in a driving environment. In the context of accident anticipation, the MDP model aims to \emph{exploit} the immediate visual cues for accident anticipation and also \emph{explore} more possibilities of accident scoring and attention allocation. 

Our proposed DRIVE model is illustrated with the MDP perspective in Fig.~\ref{fig:overview}. 
The DRIVE model simultaneously learns the policies of accident anticipation and fixation prediction based on a deep reinforcement learning (DRL) algorithm. At each time step, the agent takes actions to predict the occurrence probability of a future accident, as well as the fixation point indicating where drivers will look in the next time step. Our environmental model dynamically provides the observation state by considering both the bottom-up and top-down visual attention, which is recurrently modulated by the actions from the previous time step. We develop a novel dense anticipation reward to encourage early and accurate prediction, as well as a sparse fixation reward to enable visual explanation. Moreover, to effectively train the DRIVE model on real-world datasets, substantial improvements are made based on the DRL algorithm SAC~\cite{HaarnojaICML2018}. Our method is demonstrated to be effective on the DADA-2000 dataset~\cite{FangITSC2019}, and can be easily extended to the DAD dataset~\cite{ChanACCV2016} without fixation annotations.

The proposed approach differs from existing works~\cite{ChanACCV2016,SuzukiCVPR2018,CorcoranCRV2019, BaoMM2020} that are formulated within the supervised learning (SL) framework. The proposed DRL-based solution is fundamentally superior to SL in that the DRL could utilize immediate observations to achieve a long-term goal, i.e., making early decision for anticipating future accidents. Moreover, according to~\cite{kimECCV2018}, our method is introspectively explainable as compared to~\cite{ChanACCV2016,CorcoranCRV2019}, which simply provide rationalizations (post-hoc explanation), since we explicitly formulate drivers' visual attention during model learning. Our experimental results also validate that the learned visual attention serves as the causality of the outcome from the agent.
The main contributions are threefold: 

\begin{itemize}
    \item The DRIVE model is proposed for traffic accident anticipation from dashcam videos based on deep reinforcement learning (DRL).
    \item The DRIVE model is visually explainable by explicitly simulating the human visual attention within a unified DRL framework.
    \item The proposed dense anticipation reward and sparse fixation reward are effective in training the model by our improved DRL training algorithm.
\end{itemize}

\begin{figure*}[t]
    \centering
    \includegraphics[width=\linewidth]{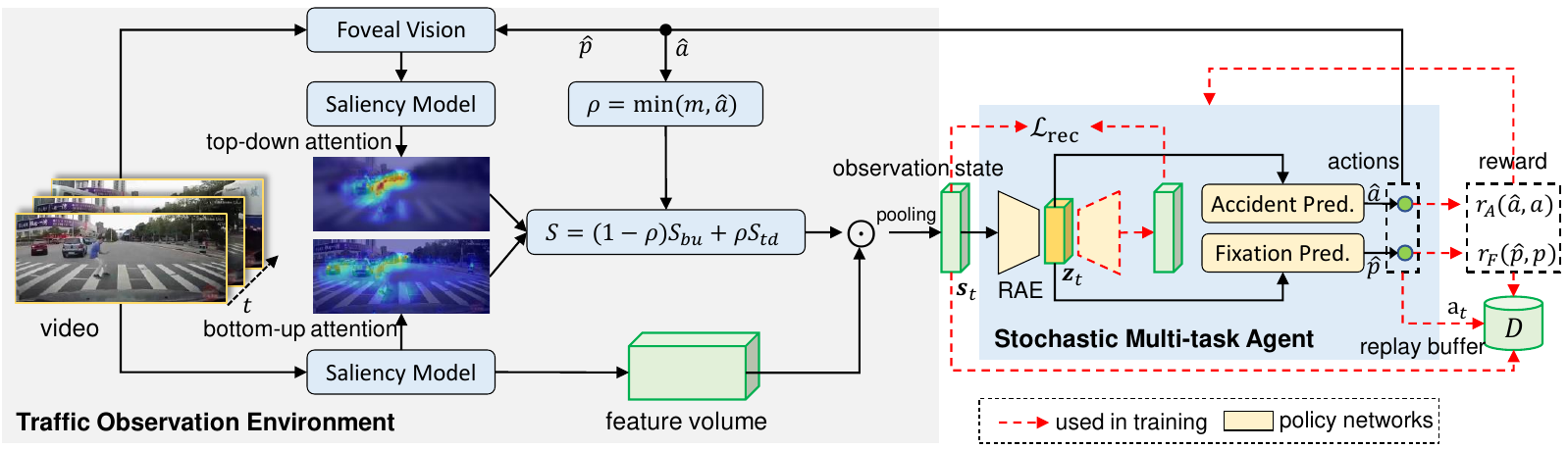}
    \captionsetup{font=small,aboveskip=5pt}
    \caption{\textbf{The DRIVE Model.} At each time step $t$, the traffic observation environment model (left part) acquires visual attention from bottom-up and top-down pathways, generating an observation state $\mathbf{s}_t$ by the dynamic attention fusion (equation box) and feature pooling. The stochastic multi-task agent (right part) takes $\mathbf{s}_t$ as input to predict the actions $\mathbf{a}_t$ which includes the accident score $\hat{a}$ and the next fixation point $\hat{p}$. All the states, actions, and rewards are collected in the replay buffer $\mathcal{D}$ to train the two policy networks of the agent. }
    \label{fig:framework}
\end{figure*}

\section{Related Work}

\textbf{Traffic Accident Anticipation.} Different from recent works on accident detection~\cite{YaoIROS2019,HuangTSAS2020}, accident recognition~\cite{YouECCV2020}, and early action/activity prediction~\cite{chenECCV2020,kongPAMI2018}, the accident anticipation problem is more challenging as the model needs to make an early decision before the accident occurs. The accident anticipation task was first formally proposed by Chan \etal~\cite{ChanACCV2016}, in which they proposed a DSA-RNN method to solve the accident anticipation problem. This method is based on object detection and dynamic soft-attention on each time step and uses LSTM to sequentially predict the accident score. In~\cite{SuzukiCVPR2018}, an adaptive loss for early accident anticipation was introduced. Based on these works, Fatima~\etal~\cite{Fatima2020} proposed a feature aggregation method for LSTM-based sequential accident score prediction. Inspired by the success of the two-stream design, Corcoran \etal~\cite{CorcoranCRV2019} adopted RGB video and optical flow for accident anticipation. Neumann~\etal~\cite{NeumannCVPRW2019} formulated the temporal accident scores as a mixture of Gaussian distribution and proposed to use 3D-CNN to predict the sufficient statistics of the distribution. Recently, Bao~\etal~\cite{BaoMM2020} proposed to use GCN and Bayesian deep learning for traffic accident anticipation. In addition to dashcam video used in these works, Shah \etal~\cite{ShahT4SW2018} utilized surveillance videos to anticipate traffic accidents. Zeng~\etal~\cite{ZengCVPR2017} recently proposed to anticipate failing accidents by localizing risky regions within an RNN framework. 

By investigating these works, we found that they typically adopted recurrent neural networks or 3D convolutional networks as the model architecture. However, their supervised learning (SL) design requires large amounts of annotated training data. In terms of explainability,~\cite{ChanACCV2016,CorcoranCRV2019} only give post-hoc bounding box explanations, which are essentially rationalizations rather than introspective explanations.

\textbf{RL-based Visual Attention.} Visual attention has been studied for several decades and it has been widely modeled as a Markov process~\cite{liechty2003,LEMEUR2015152}. The earlier work~\cite{OgnibeneBook2008} utilized the actor-critic RL algorithm on top-down attention modeling. Mnih~\etal~\cite{MnihNIPS2014} developed an RL-based recurrent visual attention model for image classification. Jiang~\etal~\cite{JiangTNNLS2016} used the Least-Squares Policy Iteration for visual fixation prediction. Recent works such as~\cite{XuPAMI2018} and~\cite{HuCVPR2017} implemented deep RL algorithms for 360$^{\circ}$ video-based human head movement prediction. In addition to RL methods, inverse reinforcement learning (IRL) algorithms take the advantages of expert demonstrations to train policy networks for task objectives, and a recent work~\cite{yangCVPR2020} showed that IRL can be leveraged to predict goal-directed human attention. Zhang~\etal~\cite{ZhangECCV8} proposed an imitation learning framework by using human fixations to learn a policy network for Atari games. In this paper, different from these works, we integrate the visual attention and the traffic accident anticipation into a unified RL framework in a real-world environment.

\textbf{Explainable Self-driving.} For self-driving applications, it is important to provide explainable decision making so that the self-driving system can be trusted by humans. Similar to our work, recently Xia~\etal~\cite{xiaWACV2020} proposed to use the foveal vision mechanism to model the human visual attention for driving speed prediction. Kim~\etal~\cite{kimICCV2017} used the visual attention model and causal filtering to visually explain the predicted steering control, i.e., steering angle and speed. Based on this work, \cite{kimECCV2018} further proposed to combine both visual attention and textual description for self-driving behavior explanation. Though there are existing works investigating the visual attention of drivers in traffic scenario~\cite{DengTITS2020,XiaACCV2018,allettoCVPRW2016}, few of them simultaneously formulate the up-stream visual attention and down-stream accident anticipation into a unified learnable model. Inspired by these works, in this paper we propose that the traffic accident anticipation can be visually explained by explicitly modeling the visual attention behavior of ego-vehicle drivers.

\section{Methodology}

\textbf{Framework Overview.} Fig.~\ref{fig:framework} illustrates the framework of the DRIVE model. Given a dashcam video as input, the stochastic multi-task agent (right part) recurrently outputs the accident score $\hat{a}$ and the next fixation $\hat{p}$ at each time step based on the observation state from the environment (left part). In particular, the environment is built by considering the bottom-up and top-down attention of the dashcam video frames, while the agent consists of a shared state auto-encoder and two parallel prediction branches. The two actions $\hat{a}$ and $\hat{p}$ are guided by the reward $r_{A}$ and $r_{F}$ respectively to encourage earliness, correctness, and attentiveness. During inference, the DRIVE model simultaneously observes the driving environment by visual attention allocation to risky regions and predicts the occurrence probability of a future accident by the trained agent.

\textbf{Problem Setup.} In this paper, we follow the task setting in the existing literature~\cite{ChanACCV2016,BaoMM2020}. A traffic accident anticipation model aims to predict a frame-level accident score $a^{t}$ that indicates the probability of the accident occurrence in the future. To evaluate the performance, \emph{Time-to-Accident} (TTA) $tta = \max (0, t_a - t)$ is used to evaluate \emph{earliness}, where $t_a$ is the actual beginning time of an accident and $t$ is the first point in time when the predicted score is higher than a threshold $a_0$, i.e., $a^{t} > a_0$. A larger $tta$ indicates the earlier time the model can anticipate the traffic accident. Besides, binary classification and saliency evaluation metrics are adopted to evaluate the \emph{correctness} and \emph{attentiveness}.

Inspired by the natural decision-making process of human drivers, i.e., observe and anticipate, we formulate the traffic accident anticipation and fixation prediction tasks as a unified Markov Decision Process (MDP). Formally, let a tuple $(\mathcal{S},\mathcal{A},P,R,L,\gamma)$ represent a discounted MDP with finite horizon (video length) $L$, where $\mathcal{S}$ and $\mathcal{A}$ are spaces of action and state, $R$ defines the reward for each state-action pair, and $\gamma \in (0,1]$ is a discount factor. In this paper, the action $\mathbf{a}_t$ in the action space $\mathcal{A}$ consists of accident score $a^{t}$ and the next fixation point $p^t=(x^{t+1},y^{t+1})^T$ defined in the image domain such that $\mathbf{a}_t=(a^{t},x^{t+1},y^{t+1})^T$. The state $\mathbf{s}_t$ is shared with the two kinds of actions. The state representation and action policy will be introduced in Sec.~\ref{sec:state} and~\ref{sec:action}, respectively. Note that $P$ defines the state transition model, i.e., $P(\mathbf{s}_{t+1}|\mathbf{s}_t, \mathbf{a}_t)$. In our method, the state transition $P$ is achieved by the fixation prediction module (Eq.~\eqref{eq:policy}) and the environment observation module (Eq.~\eqref{eqn:att} and~\ref{eqn:state}). In Sec.~\ref{sec:reward} and~\ref{sec:training}, the reward design and training algorithm will be discussed, respectively.

\subsection{Traffic Observation Environment}
\label{sec:state}

To anticipate a traffic accident, the observation state needs to be discriminative to distinguish accident-relevant cues from the cluttered traffic scene. In this paper, we are inspired by the perception mechanism of the human visual system. It is widely acknowledged that visual perception is dependent on two distinct types of attention procedure, i.e., bottom-up attention and top-down attention~\cite{Connor2004}. The bottom-up attention is determined by the salient visual stimuli from the sensory input, while the top-down attention is driven by the browsing task to achieve a long-term cognitive goal. These two mechanisms have been demonstrated to be successful in modeling the visual attention of drivers in traffic scene~\cite{dengITSC2014,dengTITS2017}. For traffic accident anticipation, observing the entire scene is inefficient while the attention mechanism can be utilized to capture the discriminative accident-relevant cues for better traffic observation state modeling.

\begin{figure}
\setlength{\abovecaptionskip}{0.5pt}
  \centering
  \hspace{0mm}
      \subcaptionbox{Full Frame\label{fig_src_im}}{
        \includegraphics[width=0.475\linewidth]{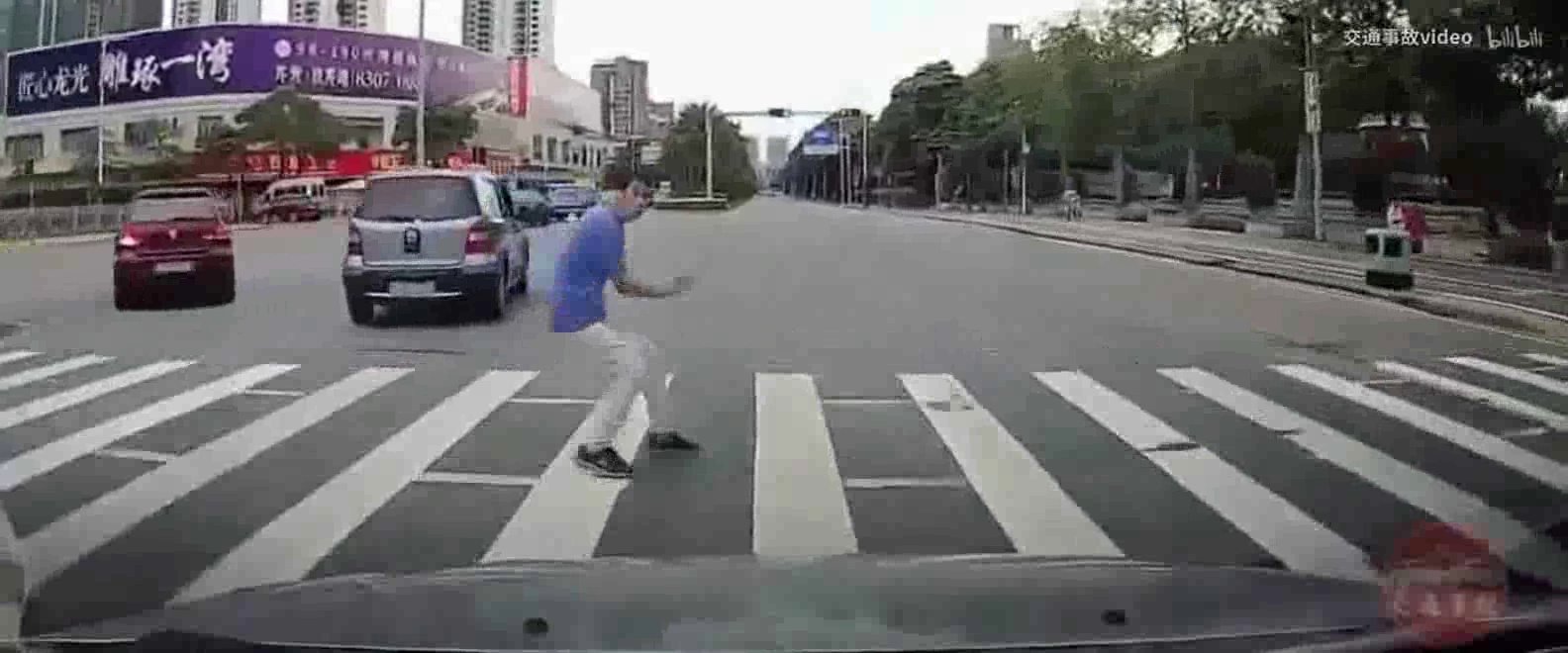}
      }%
      \subcaptionbox{Foveal Frame \label{fig_foveal_im}}{
        \includegraphics[width=0.475\linewidth]{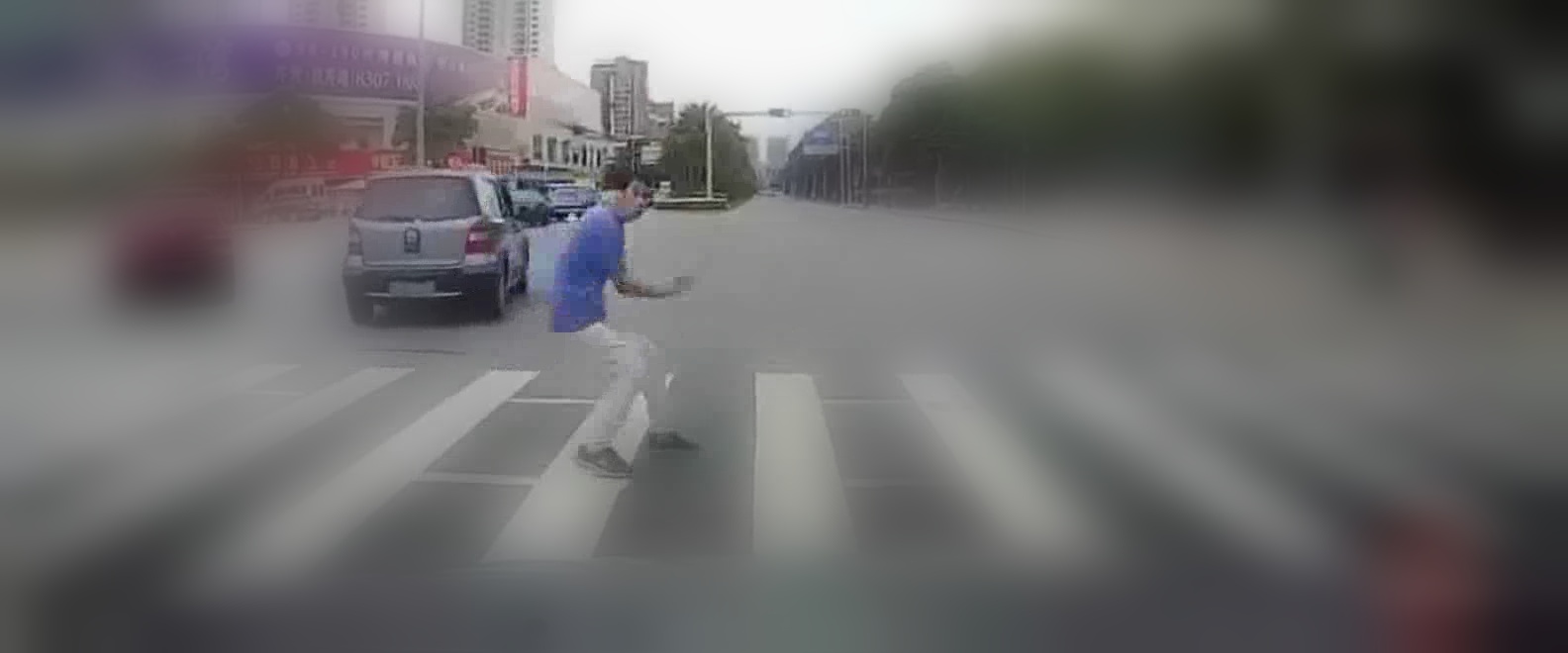}
      }%
      \vfill
      \subcaptionbox{Bottom-up Attention \label{fig_src_sal}}{
        \includegraphics[width=0.475\linewidth]{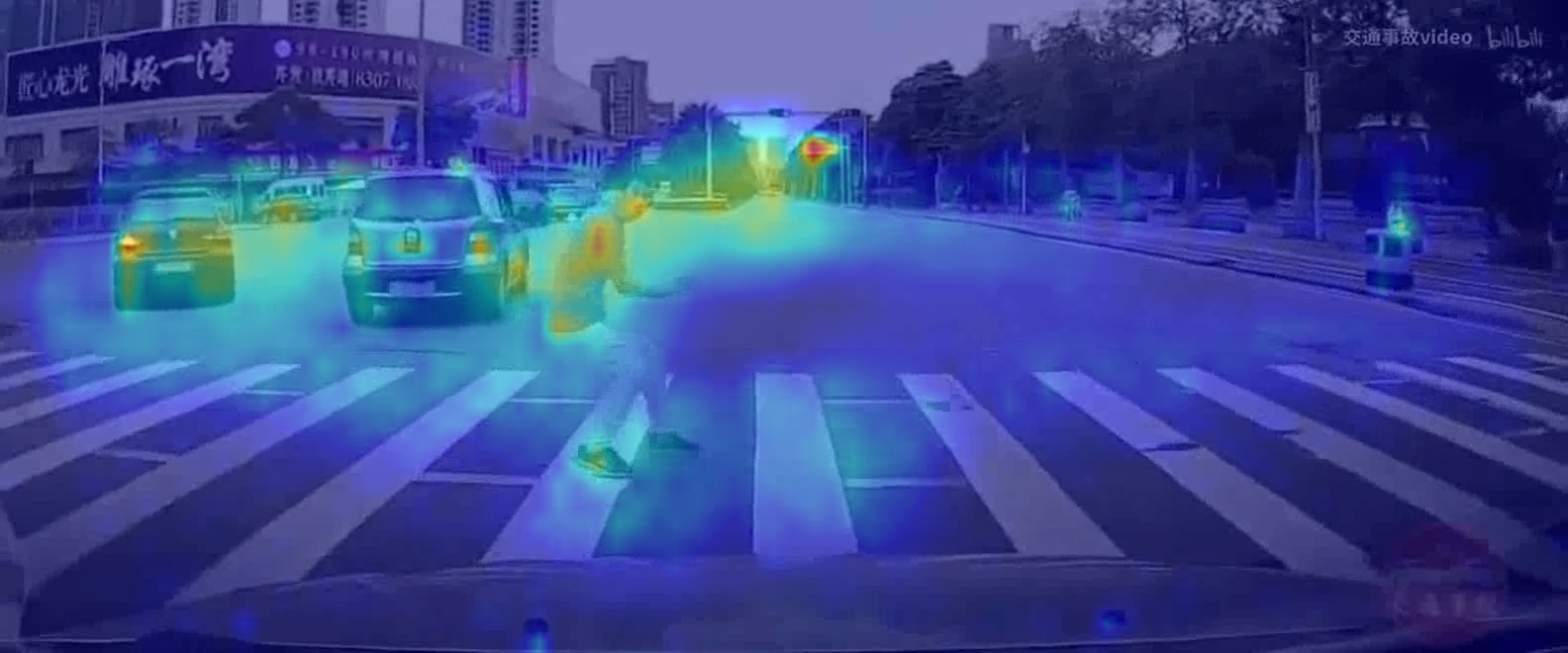}
      }%
      \subcaptionbox{Top-down Attention \label{fig_foveal_sal}}{
        \includegraphics[width=0.475\linewidth]{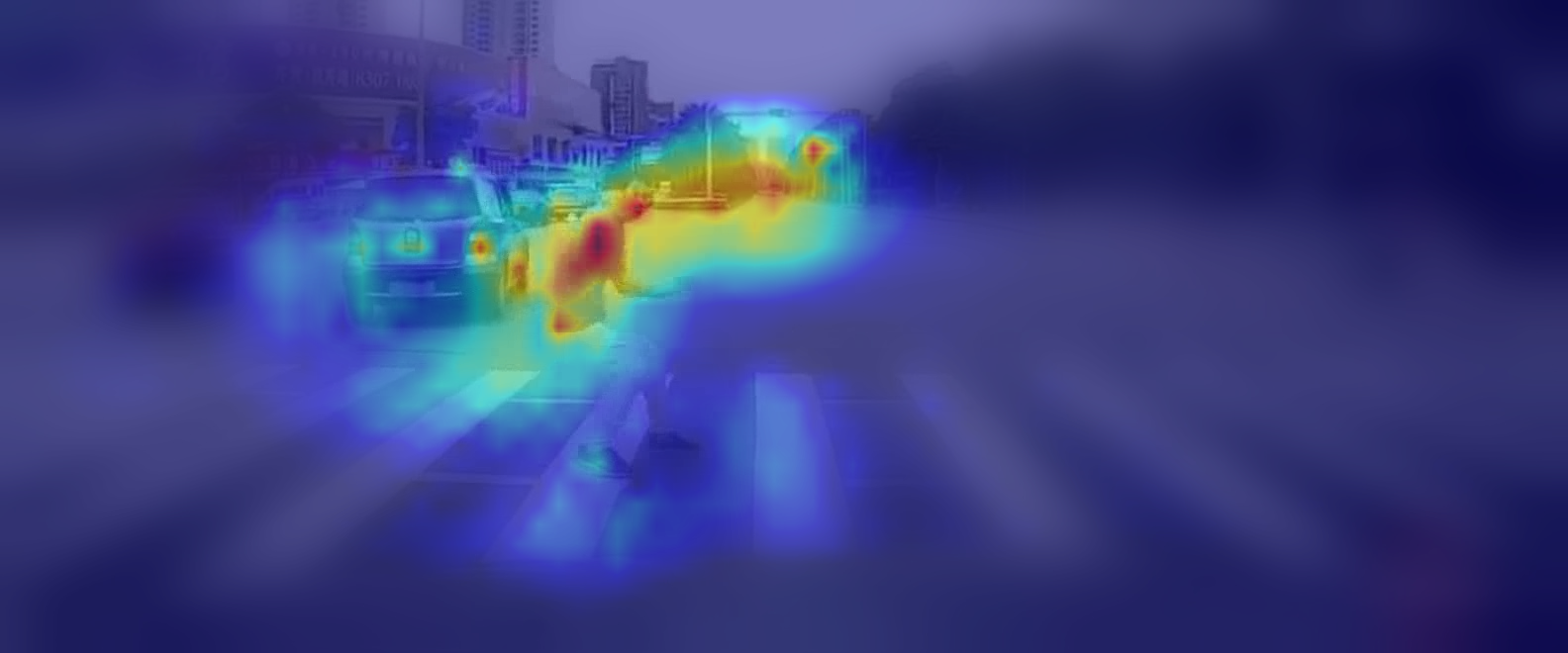}
      }%
  \captionsetup{font=small,aboveskip=3pt}
  \caption{\textbf{Examples of Foveation and Attention.} With a saliency model, the full frame $I$  (Fig.~\ref{fig_src_im}) and its foveated frame $F(I,p)$ (Fig.\ref{fig_foveal_im}) are used to generate bottom-up attention $G(I)$ (Fig.~\ref{fig_src_sal}) and top-down attention $G(F(I,p))$ (Fig.~\ref{fig_foveal_sal}), respectively.}
\label{fig:attention}
\end{figure}

\textbf{Traffic Attention Modeling.} Given the observation of current time step $I^t$, the bottom-up attention $S_{bu}^{t}\in \mathbb{R}^{H\times W}$ is simulated by a saliency prediction module $G$, i.e., $S_{bu}^{t}=G(I^t)$, where $G$ is instantiated by recent deep convolutional neural networks (CNNs) such as~\cite{CorniaICPR2016,Min2019ICCV} so that feature extraction can be shared with the saliency module. As the saliency module is not updated by the actions, $S_{bu}^{t}$ is only determined by the appearance of video frames.

To simulate top-down attention $S_{td}^{t}\in \mathbb{R}^{H\times W}$, we propose an auxiliary task to predict the fixation point $p^t\in \mathbb{R}^2$, which will dynamically guide the visual attention allocation to the risky region at each time step. Specifically, $S_{td}^{t}$ is computed by applying a foveal vision module $F$ before feeding into the saliency module, i.e., $S_{td}^{t}=G(F(I^t, p^t))$, where $F$ is implemented by the widely used method in~\cite{Geisler1998}. As $S_{td}^{t}$ is dependent on the action of fixation prediction, the agent thus dynamically interact with the attention-based observation environment. 

In this paper, both the $S_{bu}^{t}$ and $S_{td}^{t}$ are normalized in $[0, 1]$ to follow their probability nature. Fig.~\ref{fig:attention} visualizes them along with corresponding video frames. It clearly shows that bottom-up attention highlights the most salient objects while the top-down attention is more centralized in the risky region. This is because the foveal vision filters out irrelevant visual stimuli and only attends to the fixated area that indicates high risk for a future accident.

To combine the two attention mechanisms, we propose a novel dynamic attention fusion (DAF) method which is a weighted-sum of $S_{bu}^{t}$ and $S_{td}^{t}$:
\begin{equation}
    S^{t} = (1-\rho^{t})S_{bu}^{t} + \rho^{t}S_{td}^{t},
\label{eqn:att}
\end{equation}
where $\rho^{t}$ is defined as $\rho^{t}=\min(m, a^t)$. Here, $a^{t} \in [0, 1]$ is the predicted accident score and $m \in (0,1)$ serves as a hyperparameter. By introducing $m$ to clip $a^t$, instead of directly using $a^t$, the DAF gains flexibility to utilize the learned top-down attention, because $m$ controls the maximum percentage that $S_{td}^{t}$ is utilized ($\rho^t \leq m$). Note that for any $a^t < m$, we have $\rho^t=a^t$ such that $a^t$ and $1-a^t$ are used as the weighting factors. The motivation is that the more probable there will be an accident ($a^t \rightarrow 1$) at the current time step, the more confident that the predicted top-down attention can be utilized at the next time step. 


The benefits of Eq.~\eqref{eqn:att} are enormous. Because both $\rho^{t}$ and $S_{td}^{t}$ are dependent on the actions from the agent, the proposed DAF method dynamically fuses visual attention by considering both the immediate observation from the environment and the previous decision made by the agent. Our experimental results show that DAF performs better accident anticipation than the static attention fusion (SAF), i.e., manually set a fixed weighting factor. Furthermore, because the attention mechanism is explicitly formulated for accident anticipation, the resulting decisions of the agent can be visually explained by telling which region is risky.

\textbf{State Representation.} 
Since CNNs show extraordinary capability to extract appearance features, we propose to utilize the feature volume $V^t \in \mathbb{R}^{C\times H\times W}$ from CNN-based saliency model $G$ for state representation. To save the GPU memory usage while maintaining the representation capability of CNN features, the feature maps of the volume $V^t$ are aggregated and further $L_2$-normalized by global max pooling ($\Tilde{f}_{GMP}$) and global average pooling $\Tilde{f}_{GAP}$. The normalized features are then concatenated as the observation state representation:
\begin{equation}
    \mathbf{s}_t^{i} = \text{cat} \left(\Tilde{f}_{GMP}(S^{t} \odot V_i^t), \Tilde{f}_{GAP}(S^{t} \odot V_i^t)\right),
\label{eqn:state}
\end{equation}
where $\odot$ is the element-wise product on the $i$-th channel of the feature volume $V^t$, and $\text{cat}()$ is the concatenation along the channel dimension.  

\subsection{Stochastic Multi-task Agent}
\label{sec:action}

To simultaneously perform the accident anticipation and fixation prediction, the observation state $\mathbf{s}_t$ is shared with the two tasks. The state sharing brings two benefits. First, the state sharing establishes the causality relationship between the two tasks such that the visual attention modulated by the fixation prediction task could introspectively explain the accident anticipation outcomes, which distinguish our method from existing explanation-by-rationalization methods~\cite{ChanACCV2016,CorcoranCRV2019}. The causal attention is also recently studied for explainable self-driving~\cite{kimICCV2017,kimECCV2018}. 
Second, it significantly saves the communication workload between the environment and the agent, especially when the state $\mathbf{s}_t$ is of high dimensionality.

The quality of the state $\mathbf{s}_t$ is essential for improving the sample efficiency in DRL-based training. One of the typical ways is to include auxiliary observation reconstruction task along with the prediction/control task of the agent~\cite{higginsICLR2017,ghoshICLR2020}. Inspired by this, we propose to use the Regularized Auto-Encoder (RAE) to encode the state $\mathbf{s}_t$ into a more compact low-dimensional latent representation $\mathbf{z}_t$, i.e., $\mathbf{z}_t=\mathcal{E}(\mathbf{s}_t)$ where $\mathcal{E}$ is the encoder part of RAE. And the decoder of RAE is to reconstruct the observation state.

To encourage more exploration of the environment, the agent policies are designed to be stochastic in recent state-of-the-art DRL algorithms~\cite{MnihICML2016,SchulmanPPO2017,HaarnojaICML2018}. In our model, an action is associated with both the accident score and the next fixation, drawn from a Gaussian, which can be leveraged for exploration. 
Therefore, the shared latent embedding $\mathbf{z}_t$ is used to predict the mean and the variance of each action dimension for the two tasks by two parallel policy networks, respectively (see the yellow boxes in Fig.~\ref{fig:framework}). In the training stage, an action $\mathbf{a}_t$ is sampled from the predicted Gaussian distribution, i.e., $ (\mathbf{a}_t) \sim \pi_\phi(\mathbf{a}_t|\mathbf{s}_t)$ where $\phi$ is the parameterized policy network. We implement the two policy networks by two fully-connected layers with ReLU activation. Besides, similar to the recent DRL-based attention models ~\cite{MnihNIPS2014,XuPAMI2018}, a  LSTM~\cite{hochreiter1997long} is utilized after the last FC layers to capture the temporal dependency of consecutive actions. In the testing stage, the predicted means of the accident score policy $\phi_A$ and the fixation policy $\phi_F$ are concatenated as action output:
\begin{equation}
    \hat{\mathbf{a}}_t = \text{cat}\left(\phi_A\left(\mathcal{E}(\mathbf{s}_t)\right), \phi_F\left(\mathcal{E}(\mathbf{s}_t)\right) \right).
\label{eq:policy}
\end{equation}

Note that we do not directly predict the top-down attention map but instead predict the fixation point as one of the actions in $\mathbf{a}_t$. The motivation is that the attention map prediction leads to a high dimension action space which is not efficient to be learned.

\subsection{Reward Functions}
\label{sec:reward}

With the observed state $\mathbf{s}_t$ and the executed action $\mathbf{a}_t$, the agent needs a scalar reward $r$ from a driving environment to guide its learning. In this paper, we propose a dense anticipation reward $r_A$ and a sparse fixation reward $r_F$ to encourage early, accurate, and explainable decision such that the total reward at each time step is $r=r_A+r_F$.

\textbf{Dense Anticipation Reward.} For the accident score $a^t$, we propose to reward it densely (for all time steps) by considering both the \emph{correctness} and \emph{earliness} at each time step. Given a score threshold $a_0$, we propose a temporally weighted XNOR\footnote{XNOR: \url{https://en.wikipedia.org/wiki/XNOR_gate}} (also called Equivalence Gate) measurement as the accident anticipation reward $r_A$:
\begin{equation}
    r_A^t = w_t \cdot \text{XNOR} \left[\mathbb{I}[a^t > a_0], y\right],
\label{eqn:r_a}
\end{equation}
where $w_t$ is the weighting factor and  $\mathbb{I}(\cdot)$  is an indicator function. The binary label $y \in \{0,1\}$ and $y=1$ indicates there will be an accident in the future part of the video. The motivation to use XNOR is that it assigns one as the reward to the true predictions (either true positives or true negatives), while assigns zero reward to false predictions. Though in autonomous driving scenario, false negative is more detrimental than false positive, it is non-trivial to manually design the weights to achieve the balance and it is out of the scope in this paper. 

Furthermore, to encourage early anticipation (\emph{earliness}), the temporally weighting factor $w_t$ in Eq.~\eqref{eqn:r_a} is designed as a normalized expression such that $r_A$ and $r_F$ can be numerically balanced with the same magnitude scale:
\begin{equation}
    w_t = \frac{1}{e^{t_a}-1} \left(e^{\max(0, t_a-t)} - 1\right),
\label{earliness}
\end{equation}
where $t$ and $t_a$ are the current time step $t$ and the beginning time of a future accident, respectively. This factor exponentially decays from 1 to 0 before the accident occurs. Therefore, the earlier the decision is made, the larger reward will be given for the true positive prediction. After the accident occurs at $t_a$, there is no need to reward the agent.

Compared with the exponential binary-cross entropy loss in existing accident anticipation works~\cite{ChanACCV2016,BaoMM2020}, our dense anticipation reward is more appropriate for DRL training.

\textbf{Sparse Fixation Reward.} Different from rewarding the accident scores, rewarding the predicted fixations is more challenging as the ground truth fixation data are valuable and typically only sparsely provided for a few accident frames~\cite{FangITSC2019}. To this end, we resort to a sparse rewarding scheme that is widely used in dynamic programming and reinforcement learning. In particular, our sparse fixation reward is given by
\begin{equation}
    r_F^t = \mathbb{I}\left[t>t_a\right] \exp{\left(-\frac{||\hat{p}^t - p^t||^2}{\eta}\right)},
\label{eqn:r_f}
\end{equation}
where the indicator function $\mathbb{I}\left[t>t_a\right]$ zeroes out the rewards of predictions before a future accident occurs. The $\hat{p}^t$ and $p^t$ are 2-D coordinates of predicted and ground truth fixation point, respectively, defined in video frame space. The fixation points are normalized by the height and the width of the video frame for stable training. The motivation to use the radial kernel based on Euclidean distance is that the closer distance between $\hat{p}^t$ and $p^t$, the larger reward the agent will get. The  hyperparameter $\eta$ can be empirically set to ensure the same magnitude between $r_F^t$ and $r_A^t$.

\subsection{Model Training}
\label{sec:training}

To train the DRIVE model, we follow the soft actor-critic  SAC model~\cite{HaarnojaICML2018} but extend it to accommodate the accident anticipation task. SAC improves the exploration capacity of the traditional actor-critic RL through policy entropy maximization. 
Specifically, SAC aims to optimize the objective:
\begin{equation}
    \max_{\phi} \sum_{t=1}^{T} \mathbb{E}_{(\mathbf{s}_t,\mathbf{a}_t)\sim \rho_{\pi_{\phi}}}\left[r(\mathbf{s}_t,\mathbf{a}_t) + \alpha \mathcal{H}(\pi_{\phi}(\cdot|\mathbf{s}_t))\right]
\label{eqn:sac_obj}
\end{equation}
where $\alpha$ is the temperature that controls the contribution from the policy entropy $\mathcal{H}$. To achieve this objective, the actor which is the policy network, and the critic which approximates the state-action value function $Q(\mathbf{s}, \mathbf{a})$, are optimized in an interleaved way. 

In our model, as the stochastic multi-task agent gives separate entropy estimation for accident anticipation and fixation prediction, we propose to express the total entropy as the sum of the entropy from each task. Using the logarithm rule, $-\mathcal{H}(\pi_{\phi}(\cdot|\mathbf{s}_t))$ can be expressed as
\begin{equation}
    -\mathcal{H}(\pi_{\phi}(\hat{\mathbf{a}}|\mathbf{s})) = \log \left[\pi_{\phi_A}(\hat{a}|\mathbf{s})\cdot\pi_{\phi_F}(\hat{p}|\mathbf{s})\right].
\label{eqn:entropy}
\end{equation}
This enables SAC to be extended to the multi-task agent. 

\textbf{Update Critic.} For the critic network $Q_{\theta}$, it is updated by minimizing the soft Bellman residual:
\begin{equation}
    J(\theta) = \mathbb{E}\left[\left(Q_{\theta}(\mathbf{s}, \mathbf{a}) - y(r, \mathbf{s}^{\prime}, \mathbf{a})\right)^2\right],
\label{eqn:critic}
\end{equation}
where the target $y(r, \mathbf{s}^{\prime}, \mathbf{a})$ is greedily correlated with the reward $r$, the discounted soft Q-target $Q_{\bar{\theta}}$, and the entropy. Here, the $\bar{\theta}$ are parameters of  the soft Q-target network which is the delayed soft copy of the critic network. 
More details are provided in~\cite{HaarnojaICML2018} and our supplementary. 

\textbf{Update Actor.} The policy networks (actor) are updated to maximize Eq.~\eqref{eqn:sac_obj} by policy gradient method, which is equivalent to minimizing
\begin{equation}
    J_o(\phi) = \mathbb{E}\left[\alpha \log \pi_{\phi}(\hat{\mathbf{a}}|\mathbf{s}) - \min_{j=1,2} Q_{\theta_j}\left(\mathbf{s}, \hat{\mathbf{a}}\right)\right] + w_0||\phi||^2,
\label{eqn:jo}
\end{equation}
where $\hat{\mathbf{a}}\sim\pi_{\theta}(\cdot|\mathbf{s})$. The entropy term (logarithm part) is computed by Eq.~\eqref{eqn:entropy}. For the second term of expectation, the Clipped Double Q-learning~\cite{FujimotoICML2018} is used in practice. 
In this paper, we add an $L_2$ regularizer term for the policy network parameters $\phi$ to mitigate the over-fitting issue.

To accommodate SAC with multi-task policies in our model, we separately update each sub-policy network with corresponding losses $\mathcal{L}_A$ and $\mathcal{L}_F$ as regularizers:
\begin{equation}
\begin{aligned}
    & J(\phi_A) = J_o(\phi) + w_1\mathbb{E}\left[\mathcal{L}(\hat{a}^t,t_a,y)\right] \\
    & J(\phi_F) = J_o(\phi) + w_2\mathbb{E}\left[\mathbb{I}[t>t_a] d(\hat{p}^t, p^t)\right],
\label{eqn:actor}
\end{aligned}
\end{equation}
where $\phi=\{\phi_A, \phi_F\}$, and $(\hat{a}^t,\hat{p}^t,t_a,y,p^t)$ are sampled from a replay buffer $\mathcal{D}$. The distance $d(\cdot)$ defines the Euclidean distance. The accident anticipation loss $\mathcal{L}(\hat{a}^t,t_a,y)$ follows the definition in~\cite{BaoMM2020,ChanACCV2016}. The indicator function $\mathbb{I}[\cdot]$ ensures that only fixation points in accident frames can be accessed during training. Note that if $J_{o}(\phi)$ is removed, the SAC algorithm is reduced to a purely supervised learning (SL) without architectural modification.

\textbf{Update Temperature.} Recent works~\cite{haarnoja2018,wangICMLW2020} show that entropy-based RL training is brittle with respect to the temperature $\alpha$. In this paper, we follow the automatic entropy tuning~\cite{haarnoja2018} that updates $\alpha$ by minimizing 
\begin{equation}
    J(\alpha) = \mathbb{E}\left[ -\alpha \log \pi_{\phi} (\hat{\mathbf{a}}|\mathbf{s}) - \alpha \mathcal{H}_0 \right],
\label{eqn:alpha}
\end{equation}
where the negative target entropy $-\mathcal{H}_0$ is empirically set to the dimension of the action $\mathbf{a}$. In this paper, we found that $\alpha$ could be updated to zero such that the entropy (logarithm term) is hard to be optimized. To tackle this problem, we propose to clip $\alpha$ before it is updated:
\begin{equation}
    \alpha \leftarrow \max(\alpha - \lambda_{\alpha} \hat \nabla_\alpha J(\alpha), \alpha_0)
\end{equation}
where $\alpha_0$ is a small value for $\alpha$. This enables sufficient exploration of the agent during training.

\textbf{Update RAE.} The regularized auto-encoder (RAE) basically imposes $L_2$ regularizers on both the latent representation and model parameters for reconstruction learning:
\begin{equation}
    J_{RAE}(\beta) = \mathcal{L}_{rec}(\mathbf{s};\beta) + w_0 ||\beta||^2 + w_{\mathbf{s}} ||\mathbf{z}||^2,
\label{eqn:rae}
\end{equation}
where $\beta$ are decoder parameters and $\mathbf{z}$ is the encoded state representation by RAE encoder. Similar to the existing work~\cite{yarats2019improving}, the encoder parameters are updated by the critic loss $J(\theta)$ and the RAE loss $J_{RAE}(\beta)$ while the decoder parameters are only updated by $J_{RAE}(\beta)$. To enable the training on large-scale real-world videos, we reconstruct the observation state $\mathbf{s}$ rather than raw pixels as done in~\cite{yarats2019improving}. 

\textbf{Summary of Our DRL Contribution.} In this paper, the existing SAC algorithm is adapted to the real-world applications, which bridges the gap between simulation-based DRL applications and the challenging real-world tasks. Besides, for traffic accident anticipation, two novel reward functions by considering the earliness, correctness, and attentiveness are developed to guide the SAC-based model training. Moreover, to enable the multi-task learning by SAC, the proposed action entropy decomposition as well as other training techniques such as the temperature clipping and state reconstruction are empirically found useful.

\section{Experiments}

\textbf{Datasets.} Our method is evaluated on two traffic accident datasets, i.e., DADA-2000~\cite{FangITSC2019} and DAD~\cite{ChanACCV2016}. For the DADA-2000 dataset, we only use the beginning times of accidents and fixations of accident frames as ground truth. DAD is an accident dataset, in which the beginning times of accidents are fixed at the $90^{\text{th}}$ frame for positive clips. 
The video clips of the two datasets are all 5 seconds long. 

\textbf{Evaluation Protocols.} In this paper, we use the video-level Area Under ROC curve (AUC) to evaluate the anticipation correctness and the average time-to-accident (TTA) with score threshold 0.5 to evaluate the earliness. For classification metrics AUC, we only evaluate the predictions of accident frames since the output represents the occurrence probability of a future accident. To evaluate the visual attention, we adopt similarity (SIM), linear correlation coefficients (CC), and Kullback-Leibler distance (KLD). Smaller KL values indicate better performance.

\textbf{Implementation Details.} The proposed DRIVE model is implemented with the PyTorch framework. We adopt VGG-16-based MLNet~\cite{CorniaICPR2016} as the saliency module. The saliency module is pre-trained on the fixation data of the DADA-2000 training set and the parameters are kept frozen in DRIVE training. 
For the DAD dataset, as the fixations are not annotated, we remove the fixation prediction policy and top-down attention. For all datasets, the video frames are resized and zero-padded to $480\times 640$ with an equal scaling ratio. The $m$ of $\rho$ in Eq.~\eqref{eqn:att} and score threshold $a_0$ in Eq.~\eqref{eqn:r_a} are set to 0.5 by default. 
We use Adam optimizer for all gradient descent processes and train the DRIVE for 50 epochs. Other parameter settings are in the supplementary.

\begin{table}[t]
\centering
\small
\setlength{\tabcolsep}{2.0mm}
\setlength{\extrarowheight}{0.5mm}
\captionsetup{font=small,aboveskip=2pt}
\caption{Comparison with state-of-the-art methods. Best results are marked with bold fonts. AUC and TTA evaluate the correctness and earliness of accident anticipation, respectively.}
\label{tab:sota}
\small
\begin{tabular}{l|cc|cc}
\hline
\multirow{2}{*}{Methods} &\multicolumn{2}{c|}{DADA-2000~\cite{FangITSC2019}} &\multicolumn{2}{c}{DAD~\cite{ChanACCV2016}} \\
\cline{2-5}
&AUC (\%) &TTA (s) &AUC (\%) &TTA (s) \\
\hline
DSA-RNN~\cite{ChanACCV2016} & 47.19 & 3.095 & 71.57 & 1.169 \\
AdaLEA~\cite{SuzukiCVPR2018} & 55.05 & \textbf{3.890} & 58.06 & 2.228 \\
UString~\cite{BaoMM2020} & 60.19 & 3.849 & 65.96 & 0.915\\
DRIVE (ours) & \textbf{72.27} & 3.657 & \textbf{93.82} & \textbf{2.781} \\
\hline
\end{tabular}
\end{table}

\subsection{Main Results}

\input{vis}

\textbf{Baselines.} We compare the proposed DRIVE with DSA-RNN~\cite{ChanACCV2016} and UString~\cite{BaoMM2020} since their source codes are available. We also implement the accident anticipation loss function AdaLEA~\cite{SuzukiCVPR2018} on top of the DSA-RNN method (AdaLEA). Note that all methods are using VGG-16~\cite{SimonyanVGG2014} as backbone. The AUC and TTA results on DADA-2000 and DAD datasets are reported in Table~\ref{tab:sota}.

\textbf{Results for Accident Anticipation.} It shows that our DRIVE method significantly outperforms other baselines on both DADA-2000 and DAD dataset with the AUC metric. This demonstrates that our method is advantageous to accurately anticipate if a future accident will occur or not. Note that AdaLEA achieves the best TTA performance on DADA-2000 dataset, i.e., 0.23 seconds higher than our DRIVE method. The advantage of AdaLEA on TTA metric can be attributed to that during training, AdaLEA utilizes validation set to compute TTA to drive the model to make early anticipation. In contrast, we do not use validation set guidance but still achieve comparable TTA results on DADA-2000 and much better TTA on DAD datasets.

\begin{figure}[!ht]
    \centering
    \subcaptionbox{Intervention on Attention \label{fig:bar}}{
        \includegraphics[width=0.45\linewidth]{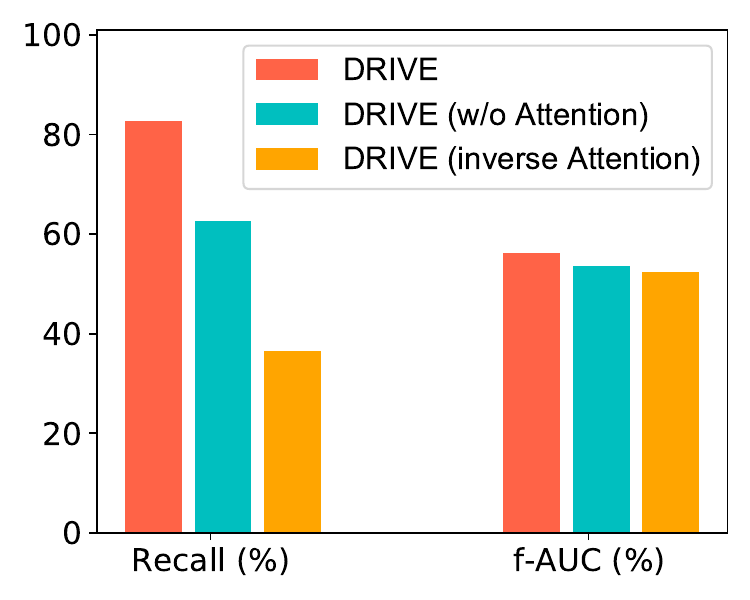}
    }
    \subcaptionbox{Reward Curves\label{fig:reward}}{
        \includegraphics[width=0.45\linewidth]{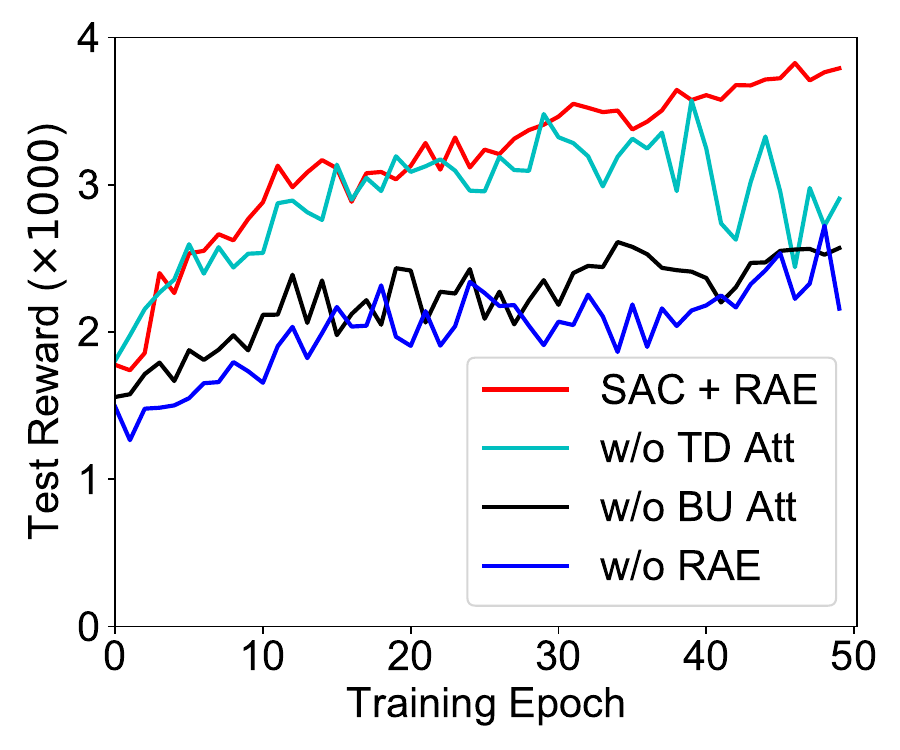}
    }
    \captionsetup{font=normalsize,aboveskip=3pt}
    \caption{Experimental results on DADA-2000 dataset.}
\end{figure}

\subsection{Visual Explanation Results}

\textbf{Correlation Results.} To investigate how the visual attention mechanism could explain the accident anticipation decision, we first jointly compare the performance of the fused saliency (Eq.~\eqref{eqn:att}) and corresponding AUC score for both the proposed dynamic attention fusion (DAF) strategy and an alternative one, i.e., static attention fusion, for which the fusion parameter is manually set without updating. Results are reported in Table~\ref{tab:sal}. We can see that DAF consistently outperforms SAF for both the saliency prediction and accident prediction (AUC), which demonstrates the superiority of DAF and the strong correlation between the visual attention and accident anticipation.

\textbf{Causality Results.} The visual attention learned by the proposed DRIVE model should exhibit the causality of accident anticipation performance. Therefore, inspired by the causal saliency analysis~\cite{kimICCV2017} and conterfactual visual explanation~\cite{GoyalICML2019}, we adopt two different intervention tests on the saliency $S^{t}$ in Eq.~\eqref{eqn:att}, i.e., removing the attention ($S^{t}\!\leftarrow\!1$) and inverse the attention ($1\!-\!S^{t}$) in testing stage. Results are reported in Fig.~\ref{fig:bar}. It shows that with either recall rate or frame-level AUC (f-AUC), the performance of the DRIVE model would decrease for both test cases, which demonstrate causality relation between the learned DAF visual attention and accident anticipation.

\textbf{Attention Visualization.} In Fig.~\ref{fig:vis}, we visualized the saliency maps on three representative time steps using test data from DADA-2000 dataset. For reference, the curve of the predicted accident probability is also presented. We can see that the bottom-up attention captures visually attentive regions while the top-down attention indicates the risky region both before and after the accident occurs. The DAF attention maps exhibit the fused attention.

\begin{table}[t]
\centering
\setlength{\extrarowheight}{0.5mm}
\captionsetup{font=small,aboveskip=2pt}
\caption{Evaluation of visual attention and accident anticipation on DADA-2000. Best results are shown in bold. The parameters (Params) represent the values of $\rho$ for SAF and $m$ for DAF.}
\label{tab:sal}
\small
\setlength{\tabcolsep}{2.8mm}
\begin{tabular}{c|c|c|ccc}
\hline
Params &Methods &AUC &SIM &CC &KLD ($\downarrow$) \\
\hline
\multirow{2}{*}{0.5} &SAF &0.645 &0.188 &0.322 &2.679 \\
&DAF &\textbf{0.659} &\textbf{0.192} &\textbf{0.331} &\textbf{2.654} \\
\hline
\multirow{2}{*}{0.8} &SAF &0.691 &0.144 &0.190 &3.087 \\
&DAF &\textbf{0.726} &\textbf{0.158} &\textbf{0.226} &\textbf{2.986} \\
\hline
\multirow{2}{*}{1.0} &SAF &0.632 &0.080 &0.079 &12.948 \\
&DAF &\textbf{0.679} &\textbf{0.112} &\textbf{0.143} &\textbf{7.836} \\
\hline
\end{tabular}
\end{table}

\subsection{Ablation Studies}

In Table~\ref{tab:ablation}, we report the results of ablation studies with DADA-2000 dataset. 
In the first row, we remove the fixation prediction policy and corresponding learning objectives, we see the AUC is about 10\% lower than our full model (the last row). The second row shows that the RAE module also contributes a lot to the performance gain. To further see if it is the DRL framework itself that leads to good performance, we keep the DRIVE architecture unchanged and only remove the $J_O(\phi)$ in Eq.~\eqref{eqn:actor} for training the multi-task policy networks such that the algorithm reduces to a supervised learning (SL). Results in the third row of Table~\ref{tab:ablation} show that DRL-based learning method (SAC) is superior to SL algorithm for accident anticipation.

To show the performance of SAC-based DRIVE variants during training, we plot their reward curves in Fig.~\ref{fig:reward}. It shows that training DRIVE model by SAC + RAE could achieve stable increasing reward. Besides, both top-down attention (w/o BU Att) and bottom-up attention (w/o TD Att) could contribute to the learning process. In particular, we see that RAE contributes most to the performance gain.

\begin{table}[t]
\centering
\setlength{\tabcolsep}{4.0mm}
\setlength{\extrarowheight}{0.5mm}
\captionsetup{font=small,aboveskip=2pt}
\caption{Ablation studies on DADA-2000 dataset. In the Type column, ``RL" and ``SL" represent reinforcement learning and supervised learning, respectively.}
\label{tab:ablation}
\small
\begin{tabular}{c|c|c|c|c}
\hline
Type & SAC &RAE &Fixations 
&AUC (\%) \\
\hline
RL & \checkmark &\checkmark & \ding{55} 
& 61.91 \\
RL & \checkmark & \ding{55} &\checkmark 
& 66.21\\
SL & \ding{55}  &\checkmark &\checkmark 
& 63.96 \\
RL & \checkmark &\checkmark &\checkmark 
& \textbf{72.27}\\
\hline
\end{tabular}
\end{table}

\section{Conclusion}

In this paper, we propose the DRIVE model to anticipate traffic accidents from dashcam videos. 
Based on deep reinforcement learning (DRL), we explicitly simulate both the bottom-up and top-down visual attention in the traffic observation environment and develop a stochastic multi-task agent to dynamically interact with the environment. The DRIVE model is learned by the improved DRL algorithm SAC. Experimental results on real-world traffic accident datasets show that our method achieves the best anticipation performance as well as good visual explainability. 

\textbf{Acknowledgement}.
This research is supported by the Office of Naval Research (ONR) grant N00014-18-1-2875 and the Army Research Office (ARO) grant W911NF-21-1-0236.
The views and conclusions contained in this document are those of the authors and should not be interpreted as representing the official policies, either expressed or implied, of the ONR, the ARO or the U.S. Government.

{\small
\bibliographystyle{ieee}
\bibliography{ref}
}

\clearpage
\input{supp}

\end{document}

%% file: vis.tex
\newcommand{\framewidth}{0.24\linewidth}

\begin{figure*}[t]
\footnotesize
\centering
\renewcommand{\tabcolsep}{0.7pt} %
\begin{tabular}{ccccc}
& Ground Truth & Bottom-Up Attention & Top-Down Attention & DAF Attention
\\
\parbox[c]{4mm}{\multirow{1}{*}[4.0em]{\rotatebox[origin=c]{90}{$t=35$}}} &
\includegraphics[width=\framewidth]{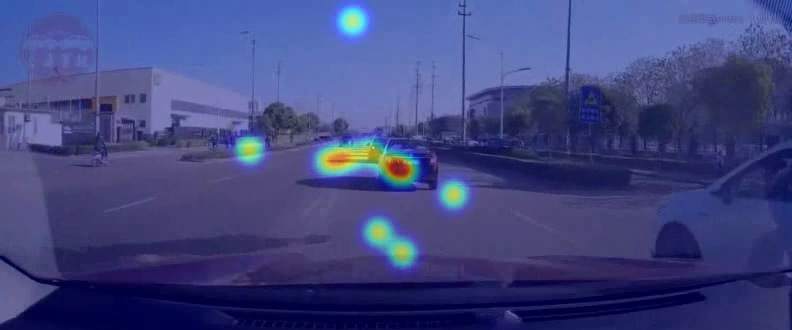} &
\includegraphics[width=\framewidth]{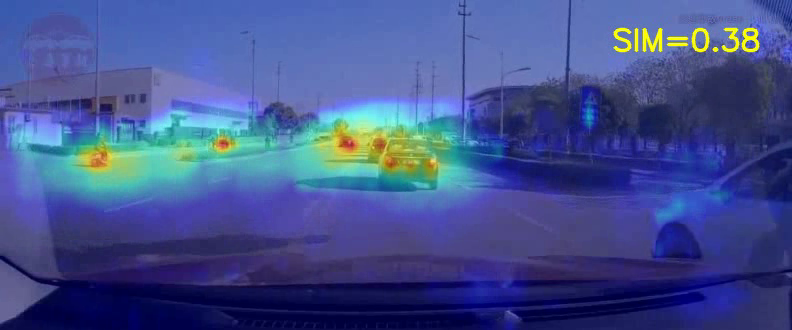} &
\includegraphics[width=\framewidth]{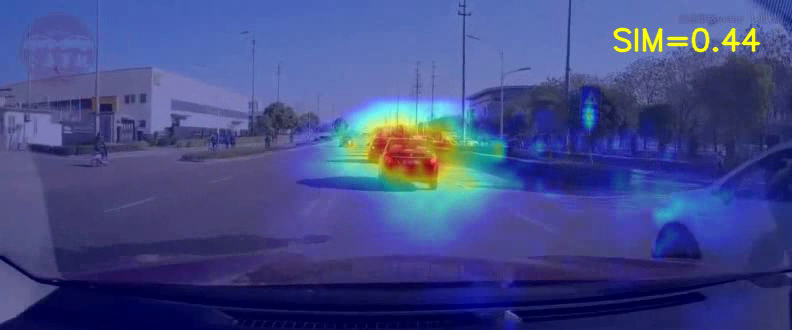} &
\includegraphics[width=\framewidth]{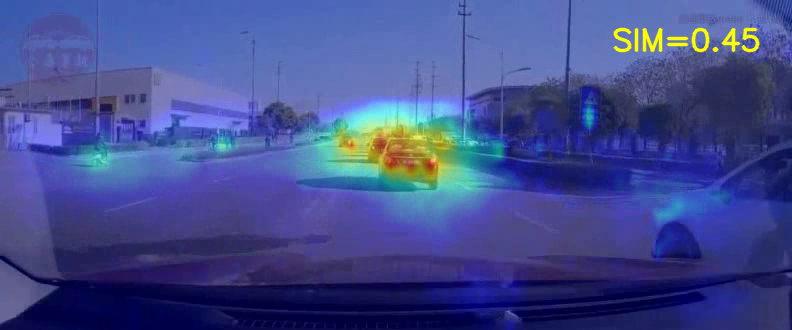}
\\
\parbox[c]{4mm}{\multirow{1}{*}[4.0em]{\rotatebox[origin=c]{90}{$t=85$}}} &
\includegraphics[width=\framewidth]{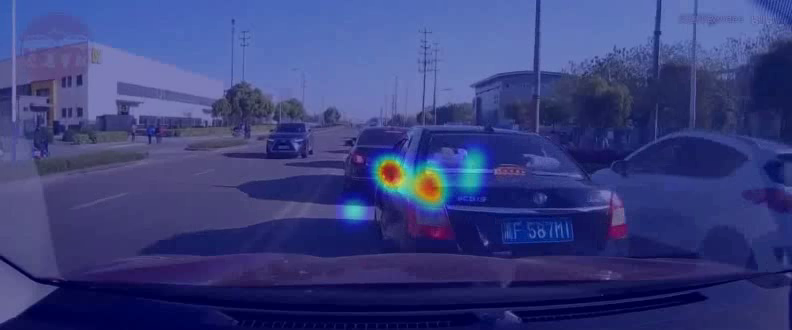} &
\includegraphics[width=\framewidth]{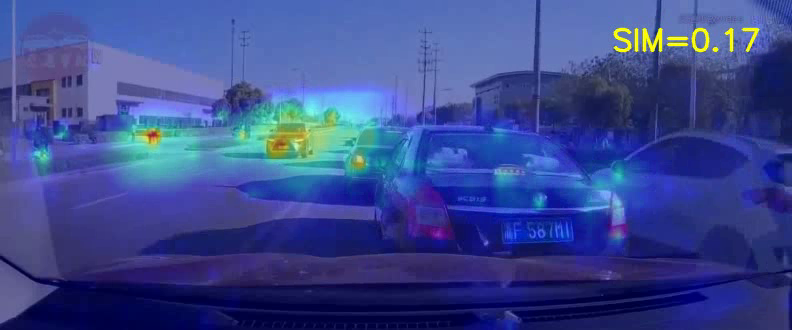} &
\includegraphics[width=\framewidth]{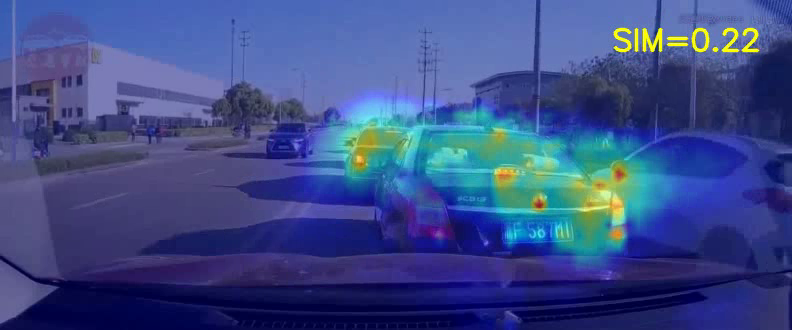} &
\includegraphics[width=\framewidth]{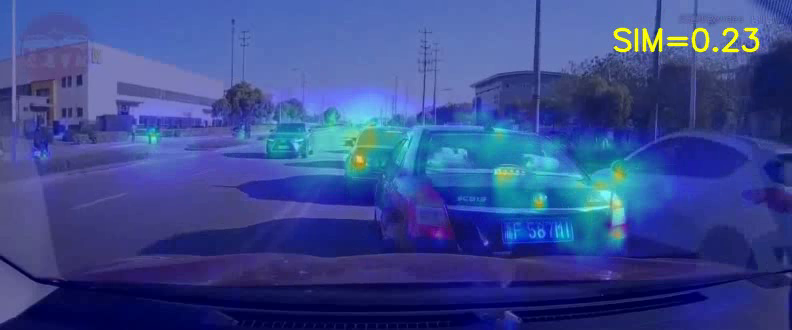}
\\
\parbox[t]{4mm}{\multirow{1}{*}[4.0em]{\rotatebox[origin=c]{90}{$t=145$}}} &
\includegraphics[width=\framewidth]{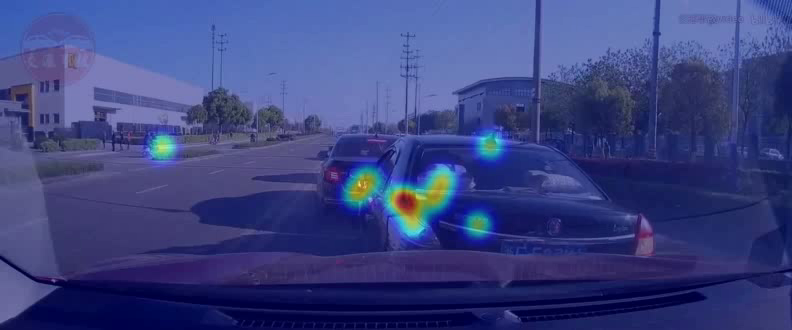} &
\includegraphics[width=\framewidth]{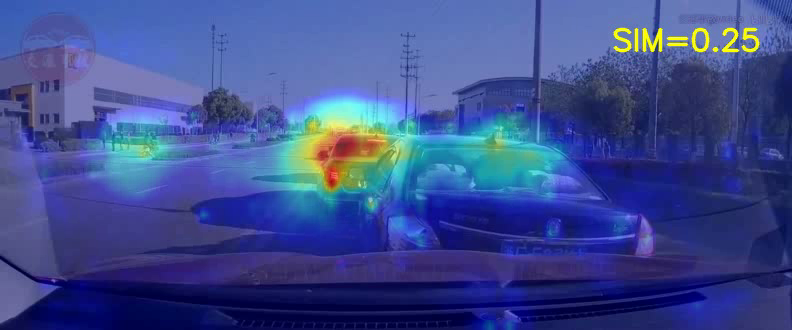}  &
\includegraphics[width=\framewidth]{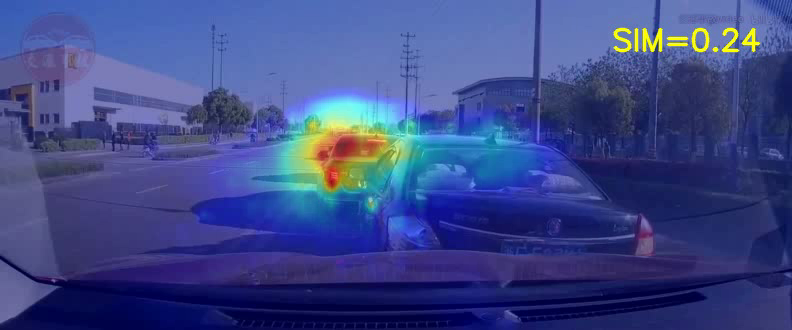} &
\includegraphics[width=\framewidth]{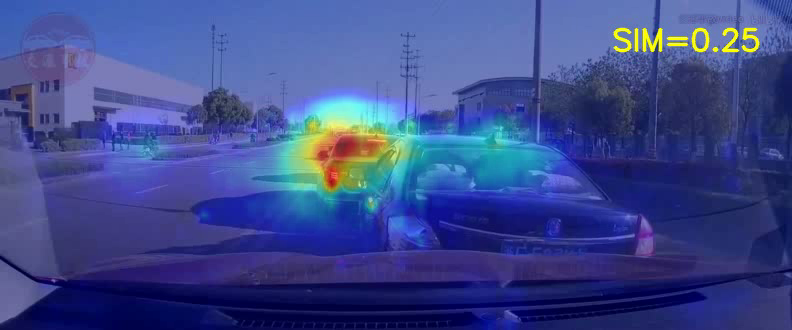} 
\\
\parbox[t]{4mm}{\multirow{1}{*}[7.0em]{\rotatebox[origin=c]{90}{Probability}}} &\multicolumn{4}{c}{\includegraphics[width=0.95\linewidth]{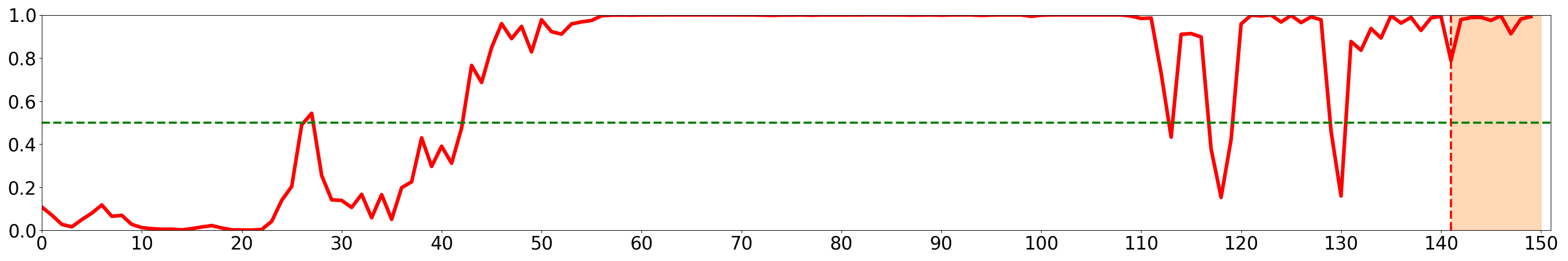}} 
\\
\end{tabular}
\captionsetup{font=small,aboveskip=3pt}
\caption{\textbf{Visualization on the DADA-2000 dataset.} The shaded region on curve figure is the accident window (FPS=30). For this example, with the operation threshold 0.5 (dashed horizontal line) and a five-frame decision window, the model could anticipate future accident at around the 42-th frame which is more than 3 seconds earlier before the accident occurs.
}
\label{fig:vis}
\vspace{-10pt}
\end{figure*}

%% file: supp.tex
\begin{appendix}

\section*{Supplementary Material}

   This document provides further details about the training algorithms of SAC~\cite{HaarnojaICML2018}, and implementation settings.

\section{Training Algorithm}

\subsection{Soft Actor Critic}
\label{sec:sac}

As introduced in Section~\ref{sec:training} in the main paper, to adapt the soft actor critic (SAC)~\cite{HaarnojaICML2018} algorithm to our DRIVE model training, original SAC algorithm needs to be substantially adapted. The Algorithm~\ref{alg:sac_detail} summarizes the training steps of the improved SAC algorithm.

At first, the transitions including the current state $\mathbf{s}_t$, action $\mathbf{a}_t$, immediate reward $r_t$, next state $\mathbf{s}_{t+1}$, and the hidden states of LSTM layer $\mathbf{h}_t$ are gathered into the replay buffer $\mathcal{D}$. For each gradient step, a mini-batch of transitions are uniformly sampled from $\mathcal{D}$ to update different model components, including the policy networks (\textbf{actor}), Q-networks (\textbf{critic}), and \textbf{RAE}. As the actor update, automatic entropy tuning, and RAE update are elaborated clearly in the main paper, here we only present more details about how the critic networks are learned during SAC training.

To update the critic, in practice, the Clipped Double Q-learning~\cite{FujimotoICML2018} is used that two identical Q-networks $\theta_i$ ($i\in \{1,2\}$) are maintained. The loss function takes the sum of the losses from the two outputs, i.e., $J(\theta)=\sum_i J(\theta_i)$, where each of them $J(\theta_i)$ is defined as the expectation of mean-squared error:
\begin{equation}
    J(\theta_i) = \mathbb{E}\left[\left(Q_{\theta_i}(\mathbf{s}, \mathbf{a}) - y(r, \mathbf{s}^{\prime}, \mathbf{a})\right)^2\right],
\end{equation}
Here, the optimization target $y(r, \mathbf{s}^{\prime}, \mathbf{a})$ is defined as
\begin{equation}
    y(r, \mathbf{s}^{\prime}, \mathbf{a}) = r + \gamma (1-d)\left(\min_{j=1,2}Q_{\bar{\theta}_j}(\mathbf{s}^{\prime}, \hat{\mathbf{a}}^{\prime}) - \alpha \log \pi_{\theta}(\hat{\mathbf{a}}^{\prime}|\mathbf{s}^{\prime})\right)
\label{eqn:q_target}
\end{equation}
where $r$ is the reward batch, $\gamma$ is the discounting factor, and $d$ labels whether the sampled transitions are at the last step $T$. Note that the sate $\mathbf{s}^{\prime}$ is the batch of next state from replay buffer, while the action $\hat{\mathbf{a}}^{\prime}$ is sampled from the output of pre-updated policy network $\pi_{\theta}$, i.e., $\hat{\mathbf{a}}^{\prime} \sim \pi_{\theta}(\cdot|\mathbf{s}^{\prime})$ which enables SAC to be an off-policy method. The entropy term $\log \pi_{\theta}(\hat{\mathbf{a}}^{\prime}|\mathbf{s}^{\prime})$ is obtained by the Eq.~\ref{eqn:entropy} in our main paper.

In this paper, the critic network parameters $\theta$ are updated more frequently than other parameters by the gradients of $J(\theta)$ to achieve more stable training. The Table~\ref{tab:hyper} summarizes the hyperparameter setting in experiments. Note that the major hyperparameters are following existing literature~\cite{HaarnojaICML2018}. For different datasets, we used the same set of hyperparameters and do not tune them specifically.

\input{alg}

\begin{table}[]
    \centering
    \setlength{\tabcolsep}{0.07\linewidth}
    \setlength{\extrarowheight}{0.5mm}
    \caption{SAC Hyperparameter Settings}
    \label{tab:hyper}
    \begin{tabular}{c|c}
    \hline
        Parameters & values \\
    \hline
        general learning rate ($\lambda$) & $3\cdot 10^{-4}$ \\
        temperature learning rate ($\lambda_{\alpha}$) & $5\cdot 10^{-5}$ \\
        discounting factor ($\gamma$) & 0.99 \\
        replay buffer size ($\mathcal{D}$) & $10^{6}$ \\
        target smoothing coefficient ($\tau$) & 0.005 \\
        temperature threshold ($\alpha_0$) & $10^{-4}$ \\
        weight decay ($w_0$) & $10^{-5}$ \\
        anticipation loss coefficient ($w_1$) & 1 \\
        fixation loss coefficient ($w_2$) & 10 \\
        latent regularizer coefficient ($w_s$) & $10^{-4}$ \\
        sparse fixation reward parameter ($\eta$) & 0.1 \\
        gradient updates per time step & 4 \\
        actor gradient updates per time step & 2 \\
        dim. of FC/LSTM layers output & 64 \\
        dim. of latent embedding ($\mathbf{z}$) & 64 \\
        dim. of state ($\mathbf{s}$) & 128 \\
        dim. of action ($\mathbf{a}$) & 3 \\
        sampling batch size & 64 \\
        video batch size & 5 \\
    \hline
    \end{tabular}
\end{table}


\section{Implementation Details} 

\begin{figure}
    \centering
    \includegraphics[width=\linewidth]{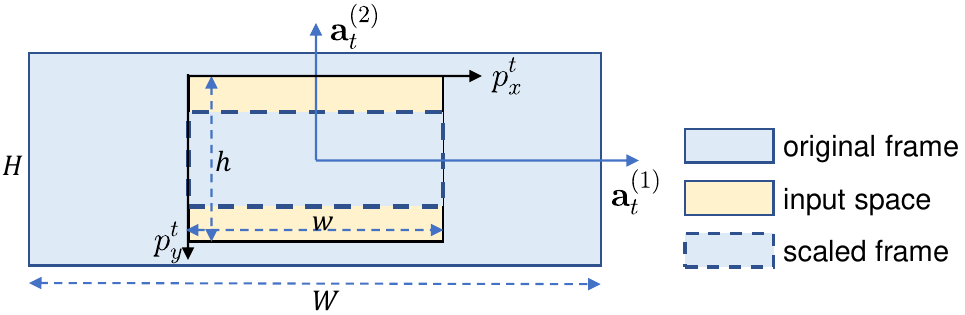}
    \caption{\textbf{The scaling process ($\psi$)}. The continuous values $\mathbf{a}_t^{(1)}$ and $\mathbf{a}_t^{(2)}$ which are within $(-1,1)$ defined in video frame space $H\times W$ are mapped into the discrete input space $h\times w$ to represent the 2-D coordinates of a fixation point.}
    \label{fig:ptscale}
\end{figure}

\textbf{Network Architecture.} As shown in Fig.~\ref{fig:framework} in the main paper, the saliency model is implemented with the existing CNN-based saliency model~\cite{CorniaICPR2016}, which takes as input with the size $480 \times 640 \times 3$ and output the feature volume $V^t$ with the size $60 \times 80 \times 64$ by default. The stochastic multi-task agent consists of a shared RAE and two policy networks, i.e., accident prediction and fixation prediction branches. In our implementation, the encoder of RAE consists of three fully-connected (FC) and the decoder is symmetric to the encoder. Each policy branch consists of two FC layers and one LSTM layer, followed by an FC layer for predicting means and an FC layer for predicting the variance. ReLU activations are used for all layers except for the last FC output layer. According to the default SAC setting, the output of policy networks $\mathbf{a}_t$ are activated by \emph{tanh} functions so that the values are constrained in $(-1,1)$. In order to map the values to accident scores $a^t$ and fixation coordinates $p^t$, we linearly scale the values by 
\begin{align}
    a^t & =0.5(\mathbf{a}_t^{(0)} + 1.0) \\
    p^t & =\psi(\mathbf{a}_t^{(1)}, \mathbf{a}_t^{(2)})
\end{align}
where the equation of $a^t$ applied to \emph{tanh} activation is equivalent to \emph{sigmoid} activation on FC layer output. The function $\psi$ maps the scaling factors (within $(-1,1)$) defined in image space $H\times W$ to the input space  $h\times w$. This scaling process is illustrated in Fig.~\ref{fig:ptscale}.

\balance

\textbf{Implementation.} Our training algorithm is implemented based on the SAC source code\footnote{\url{https://github.com/pranz24/pytorch-soft-actor-critic}}. Since the image foveation method~\cite{Geisler1998} incurs computational cost due to the  Gaussian pyramid filtering, we implement this algorithm as well as all the DRL environmental components by PyTorch to support for GPU acceleration. For DADA-2000 videos, the positive video clips (contains accident) are obtained by trimming the video into be 5 seconds where the beginning times are placed in the last one second with random jittering, while the negative video clips are randomly sampled without overlap with positive clips. The spatial and temporal resolutions for DADA-2000 videos are reduced with ratio 0.5 and interval 5, respectively, so that 30 time steps are utilized and for each step the observation frames are with the size $330\times 792$. For DAD dataset, we only reduce the temporal resolution with interval 4 so that 25 time steps of each 5-seconds video clip are used.

\end{appendix}

%% file: alg.tex
\begin{algorithm}[tb]
\caption{Improved SAC for the DRIVE Model Training}
\label{alg:sac_detail}
\begin{algorithmic}[1]
\Require $\theta_1$, $\theta_2$, $\phi$, $\beta$ \Comment{Initial parameters}
\State $\bar \theta_1 \leftarrow \theta_1$, $\bar \theta_2 \leftarrow \theta_2$ \Comment{Initialize target networks}
\State $\mathcal{D}\leftarrow\emptyset$, $\mathbf{h}_0 \leftarrow \mathbf{0}$  \Comment{Replay buffer and hidden states}
\For{each iteration}
	\For{each environment step}  
	    \State Sample actions $(\mathbf{a}_t, \mathbf{h}_t) \sim \pi_\phi(\mathbf{a}_t|\mathbf{s}_t, \mathbf{h}_{t-1})$ 
	    \State Compute state $\mathbf{s}_t$ with actions \Comment{See Eq.~\ref{eqn:state}}
	    \State Compute reward $r_t=r_A^t + r_F^t$
	    \Comment{See Eq.~\ref{eqn:r_a}-\ref{eqn:r_f}}
	    \State $\mathcal{D} \leftarrow \mathcal{D} \cup \left\{(\mathbf{s}_{t}, \mathbf{a}_{t}, r_{t}, \mathbf{h}_t, \mathbf{s}_{t+1})\right\}$ 
	\EndFor
	\For{each gradient step}   
        %
            \For{each critic update}  
    	    \State $\theta \leftarrow \theta - \lambda \hat \nabla_{\theta} J_Q(\theta)$  \Comment{Update by Eq.~\ref{eqn:critic}}
        	\EndFor
	    \State $\phi \leftarrow \phi - \lambda \hat \nabla_\phi J_\pi(\phi)$ \Comment{Update by Eq.~\ref{eqn:actor}}
	        \State $\alpha \leftarrow \max(\alpha - \lambda_{\alpha} \hat \nabla_\alpha J(\alpha)$, $\alpha_0$) \Comment{See Eq.~\ref{eqn:alpha}}
	    \State $\bar\theta\leftarrow \tau \theta + (1-\tau)\bar\theta$ \Comment{Update Q-target}
	        \State $\beta \leftarrow \beta - \lambda \hat \nabla_\beta J_\text{RAE}(\beta)$  \Comment{Update by Eq.~\ref{eqn:rae}}
	\EndFor
\EndFor
\Ensure $\theta_1$, $\theta_2$, $\phi$, $\beta$ 
\end{algorithmic}
\end{algorithm}